\documentclass[10pt,twocolumn,letterpaper]{article}

\usepackage{cvpr}              
\usepackage{multirow}
\usepackage{tikz}
\usepackage{bm}
\usepackage{graphicx}
\usepackage{amsmath}
\usepackage{amssymb}
\usepackage{booktabs}
\usepackage{cuted}
\usepackage{caption}
\usepackage{array}
\usepackage{graphicx}
\usetikzlibrary{arrows.meta,positioning,fit}
\definecolor{cvprblue}{rgb}{0.21,0.49,0.74}
\usepackage[pagebackref,breaklinks,colorlinks,allcolors=cvprblue]{hyperref}

\def\paperID{} 
\def\confName{CVPR}
\def\confYear{2026}

\title{Disentangle-then-Align: Non-Iterative Hybrid Multimodal Image Registration via Cross-Scale Feature Disentanglement}
\author{
Chunlei Zhang$^{1}$ \quad
Jiahao Xia$^{1}$ \quad
Yun Xiao$^{2}$ \quad
Bo Jiang$^{3}$ \quad
Jian Zhang$^{1}$\thanks{Corresponding author.}\\
$^{1}$Faculty of Engineering and IT, University of Technology Sydney, Australia\\
$^{2}$School of Artificial Intelligence, Anhui University, Hefei, China\\
$^{3}$School of Computer Science and Technology, Anhui University, Hefei, China\\
\texttt{\small
Chunlei.Zhang@student.uts.edu.au,\;
\{Jiahao.Xia-1, Jian.Zhang\}@uts.edu.au,\; 
} \\
\texttt{\small
\{xiaoyun, jiangbo\}@ahu.edu.cn
}
}
\begin{document}
\maketitle
\begin{abstract} 
Multimodal image registration is a fundamental task and a prerequisite for downstream cross-modal analysis. 
Despite recent progress in shared feature extraction and multi-scale architectures, two key limitations remain. First, some methods use disentanglement to learn shared features but mainly regularize the shared part, allowing modality-private cues to leak into the shared space. Second, most multi-scale frameworks support only a single transformation type, limiting their applicability when global misalignment and local deformation coexist.
To address these issues, we formulate hybrid multimodal registration as jointly learning a stable shared feature space and a unified hybrid transformation. Based on this view, we propose \textbf{HRNet}, a \textbf{H}ybrid \textbf{R}egistration \textbf{N}etwork that couples representation disentanglement with hybrid parameter prediction. A shared backbone with Modality-Specific Batch Normalization (MSBN) extracts multi-scale features, while a Cross-scale Disentanglement and Adaptive Projection (CDAP) module suppresses modality-private cues and projects shared features into a stable subspace for matching. Built on this shared space, a Hybrid Parameter Prediction Module (HPPM) performs non-iterative coarse-to-fine estimation of global rigid parameters and deformation fields, which are fused into a coherent deformation field. Extensive experiments on four multimodal datasets demonstrate state-of-the-art performance on rigid and non-rigid registration tasks. The code is available at the project website\footnote{\url{https://github.com/Chunlei0913/HRNet}}.
\end{abstract}
    
\section{Introduction}
\label{sec:intro}
With advances in sensing technologies, multimodal imaging has become increasingly widespread. However, viewpoint differences introduce geometric misalignment that defines the objective of image registration, while different imaging mechanisms induce substantial cross-modal appearance gaps, making alignment more challenging. Therefore, accurate multimodal registration is crucial for robust data fusion and for improving the reliability of downstream analysis and decision making.
\begin{figure}[tbp!]
\centering
\includegraphics[scale=0.5]{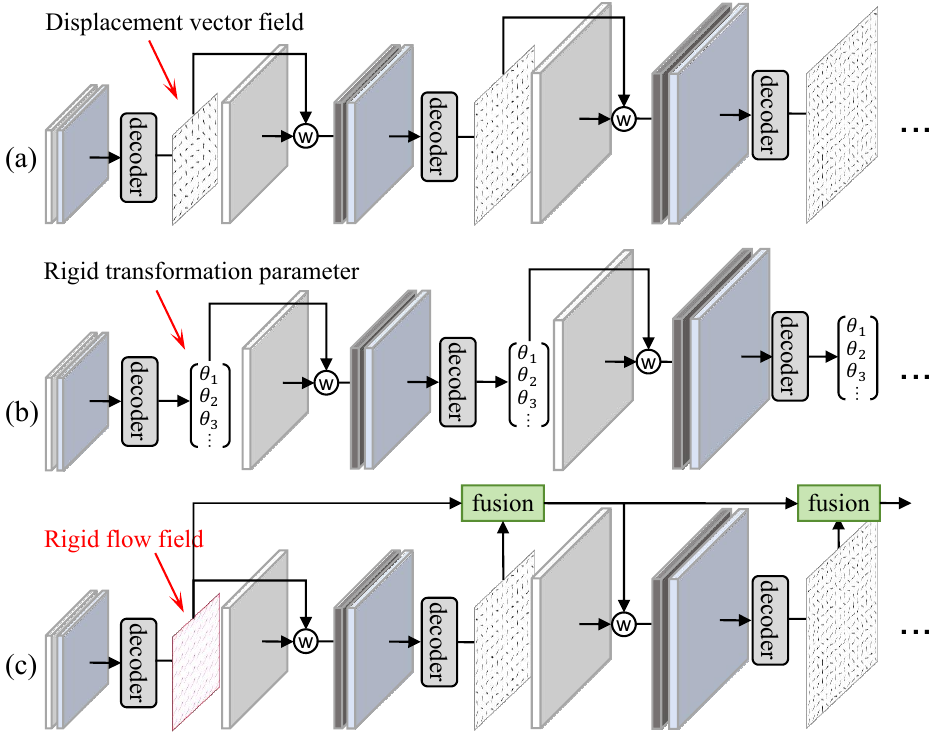}
\caption{Comparison of model structures. (a) non-rigid only, (b) rigid only; (c) Our hybrid structure.}
\label{mov1}
\end{figure}
   
\begin{figure}[tbp!]
\centering
\includegraphics[scale=0.3]{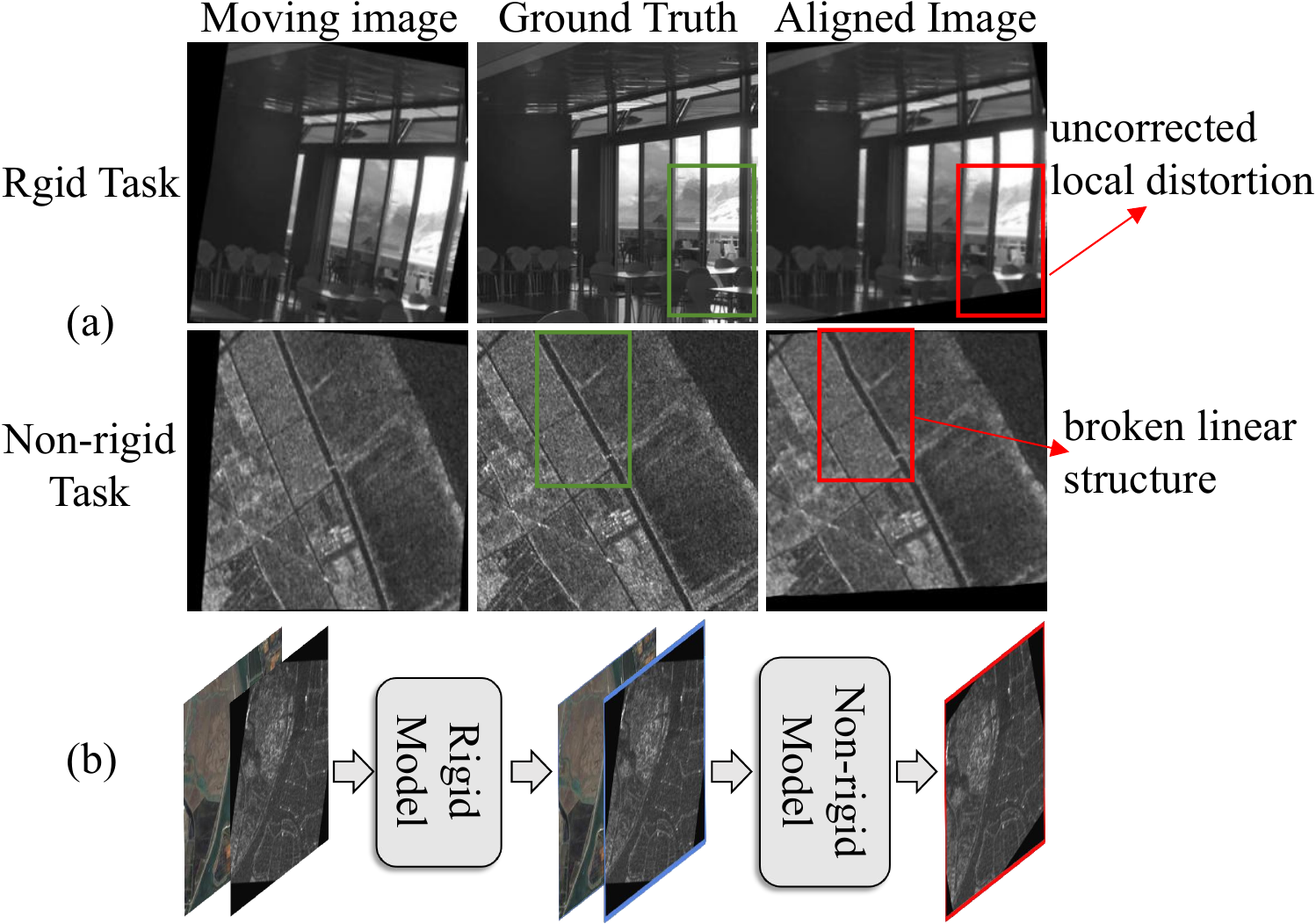}
\caption{(a) Rigid registration fails to handle local deformations (first row). Non-rigid registration may distort structural integrity under large global offsets (second row). (b) Serial hybrid registration.}
\label{mov2}
\end{figure}

Image registration has been a long-standing research problem~\cite{lowe2004distinctive,bay2006surf,li2019rift,zhang2023optical,xiao2024adrnet,li2020remote}.
Recently, deep learning–based approaches have been widely adopted for end-to-end registration, and many of them employ multi-scale strategies to improve accuracy~\cite{ye2022multiscale, zhu2024mcnet, wang2025multi}. 
However, as illustrated in Fig.~\ref{mov1}, the design of these models is typically limited to a single type of transformation, which restricts their ability to handle complex real-world scenarios involving both global misalignment and local structural variation.
This is because rigid transformations can correct global misalignment but cannot handle local deformations, whereas non-rigid transformations adapt to local variations but perform poorly under large global offsets and may distort structural integrity, as illustrated in Fig.~\ref{mov2}(a).
To leverage both, hybrid registration has been explored \cite{xiao2024adrnet,tang2020admir,gao2021deepasdm}.
However, most existing methods implement hybrid registration as a serial cascade of rigid and non-rigid stages, as illustrated in Fig.~\ref{mov2}(b). This stage-wise structure leads to two critical issues. First, rigid and non-rigid transformations are estimated in separate representation spaces, making it hard to coordinate the stages and obtain a unified hybrid deformation. Second, weak inter-stage coupling lets later predictions inherit earlier biases.
Beyond multi-scale designs, several studies seek to improve cross-modal registration by learning shared representations~\cite{hu2025ardmr,shi2023unsupervised}. Although these approaches mitigate cross-modal distribution gaps to some extent, their constraints mainly act on the shared component. They lack explicit constraints on the boundary and independence of the modality-private subspace, so private information can still leak back into the shared space and interfere with geometric correspondence.

To address these limitations, we propose HRNet, a unified \textbf{H}ybrid \textbf{R}egistration \textbf{N}etwork that serves as a general template for jointly modeling global rigid alignment and local non-rigid deformation within a single framework. HRNet consists of three components: a shared encoder, a Cross-scale Disentanglement and Adaptive Projection (CDAP) module, and a Hybrid Parameter Prediction Module (HPPM). Unlike serial schemes, HPPM performs hybrid registration in a non-iterative, coarse-to-fine manner within the shared feature space: it jointly regresses global rigid parameters and multi-scale displacement fields, encodes the rigid parameters into a global flow, and fuses this flow with progressively refined local deformation flows within a single pipeline, ultimately producing a single coherent deformation field.
To ensure that hybrid parameter prediction is always carried out in the same feature space, and to avoid rigid and non-rigid components being estimated in separate representation spaces that cannot form a unified hybrid deformation, we adopt an extract-then-disentangle strategy. Specifically, we first use a shared backbone with Modality-Specific Batch Normalization (MSBN) to extract multi-scale features, which alleviates cross-modal statistical shift and allows the backbone to focus more on geometric commonalities. On top of this, we introduce the Cross-scale Disentanglement and Adaptive Projection (CDAP) module, which follows a decompose–gate–project pipeline to learn cleaner shared representations for hybrid registration: at each scale, features are decomposed into shared and private components, cross-scale gating is used to suppress the leakage of modality-private information into the shared space, and the shared features are adaptively projected into a more stable latent subspace. At the same time, we introduce structured regularization to reduce shared–private coupling, avoid subspace redundancy, and maintain cross-scale semantic consistency. Built upon this shared space, HPPM leverages a Mamba state-space architecture to model long-range dependencies with low computational cost, and completes joint estimation of rigid and non-rigid transformations within the unified framework.

The following are our main contributions:

\begin{itemize}

\item We propose HRNet, a unified hybrid registration framework that couples feature disentanglement with hybrid parameter prediction, and achieves state-of-the-art performance on four multimodal datasets.

\item We develop a Hybrid Parameter Prediction Module (HPPM) that jointly predicts rigid and non-rigid transformations within a single coarse-to-fine pipeline, producing a unified coherent deformation field.

\item We design the Cross-scale Disentanglement and Adaptive Projection (CDAP) module, which follows a decompose–gate–project pipeline to learn multi-scale shared representations that are robust to modality-private leakage.

\item We introduce three structured regularizations, Cross-Covariance Decorrelation, Basis Orthogonality, and Cross-Scale Directional Consistency, to shape the shared feature space by reducing shared–private coupling, alleviating subspace redundancy, and enforcing cross-scale semantic consistency.
\end{itemize}

\section{Related Work}
\label{sec:Related_Work}

This section briefly reviews previous multimodal image registration methods.
To cope with the inherent modality gap, existing approaches can be broadly grouped into three representative directions.

\noindent\textbf{Multi-scale optimization.}
These approaches estimate transformation parameters in a coarse-to-fine manner, correcting global misalignment at coarser scales and refining details at finer scales.
Ye et al.~\cite{ye2022multiscale} proposed a three-stage framework that constructs a multi-scale pyramid via downsampling and accumulates transformations from low to high resolution.
Building on this idea, Zhu et al.~\cite{zhu2024mcnet} introduced iterative refinement at each scale to further improve accuracy.
For non-rigid tasks, Xu et al.~\cite{xu2022nbr} designed a bidirectional registration network with an inverse-consistency constraint to suppress one-way distortion during multi-stage refinement.
To exploit the complementary strengths of rigid and non-rigid models, hybrid registration~\cite{xiao2024adrnet,tang2020admir,gao2021deepasdm} has also been investigated. Xiao et al.~\cite{xiao2024adrnet} adopt a global-to-local two-stage pipeline, where an affine stage removes large misalignment followed by a deformable field for local correction.
However, serially concatenated hybrids remain sensitive to early-stage biases and accumulate warping artifacts, motivating a unified, single-pass hybrid formulation.

\noindent\textbf{Learning modality-shared representations.}
This line seeks to mitigate cross-modal discrepancies by learning shared features, often via disentangling shared (geometric) from modality-private (appearance) components.
Shi et al.~\cite{shi2023unsupervised} separated images into content and style features using domain adaptation and used the content features for parameter prediction.
Deng et al.~\cite{deng2023interpretable} modeled multimodal registration as disentangled convolutional sparse coding, extracting registration-related features to improve accuracy and efficiency.
In this work, we emphasize scale-aware consistency across the feature pyramid, limit private-to-shared leakage during cross-scale processing, and stabilize the shared subspace, yielding cleaner representations and more reliable correspondences under large modality gaps.

\noindent\textbf{Modality translation.}
Another line translates one modality into another to form pseudo mono-modal pairs for registration.
Wang et al.~\cite{wang2022unsupervised} and Wei et al.~\cite{wei2019synthesis} utilized CycleGAN~\cite{zhu2017unpaired}-based translation network to generate pseudo mono-modal pairs for registration.
Arar et al.~\cite{arar2020unsupervised} alternate training translation and registration under a geometry-preserving constraint, enabling the registration network to learn smooth and accurate spatial transformations with mono-modal metrics.
However, translation can introduce artificial noise~\cite{deng2023interpretable}, yielding biased correspondences; we therefore avoid synthesis and instead pursue feature disentanglement with scale-aware consistency and a stabilized shared space.

\section{Method}
\begin{figure*}[tbp!]
\centering
\includegraphics[scale=1.24]{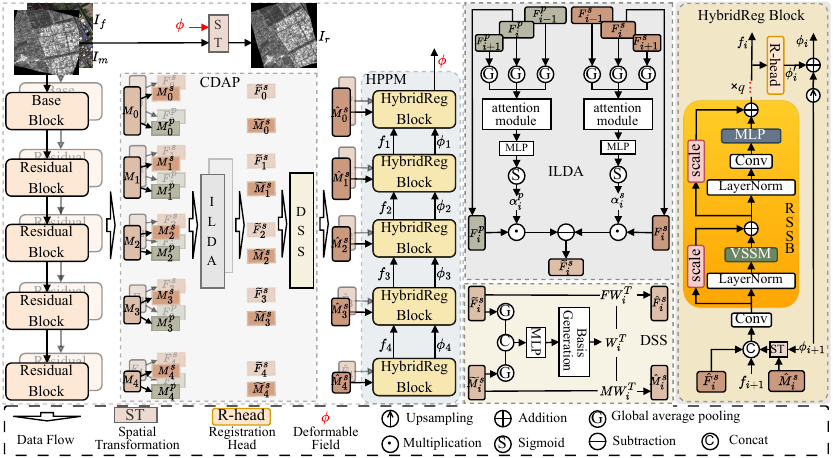}
\caption{The schematic diagram and detailed architectures of the \textbf{H}ybrid \textbf{R}egistration \textbf{N}etwork, namely HRNet, which consists of three main components: a shared backbone with Modality-Specific Batch Normalization (MSBN), a Cross-scale Disentanglement and Adaptive Projection (CDAP) module for feature disentanglement, and a Hybrid Parameter Prediction Module (HPPM) for transformation parameter estimation.}
\label{framework}
\end{figure*}

\subsection{Overview}
The overall framework of HRNet is illustrated in Fig.~\ref{framework}.
Given a fixed image $I_f$ and a moving image $I_m$, the HRNet pipeline consists of three main stages. First, the image pair $(I_f, I_m)$ is processed by a shared backbone to extract multi-scale features $F$ and $M$. Second, $F$ and $M$ are fed into the Cross-scale Disentanglement and Adaptive Projection (CDAP) module to obtain multi-scale shared features $(\hat F^s, \hat M^s)$. Finally, $\hat F^s$ and $\hat M^s$ are passed into the Hybrid Parameter Prediction Module (HPPM) to estimate the transformation $\phi$, which is then used to warp $I_m$ and produce the aligned image $I_r$  in a single forward pass.
\subsection{Shared Backbone with MSBN}
According to work~\cite{chang2019domain}, when weights are shared across modalities, a single BN stream must absorb heterogeneous statistics from different modalities, which tends to mix modality-private information into the shared space and degrades registration performance.
To mitigate this, we adopt a shared backbone with Modality-Specific Batch Normalization (MSBN): convolutional weights are shared to learn geometric commonalities, while BN means, variances, and affine parameters are maintained per modality, allowing normalization to absorb appearance and radiometric discrepancies.
This design introduces negligible overhead while producing cleaner modality-normalized features.
Formally,
    \begin{equation}
    \begin{split}
            F &= \mathrm{Backbone}(I_{f}) , F \in \{F_0,F_1,...,F_i\},\\
            M &= \mathrm{Backbone}(I_{m}) , M \in \{M_0,M_1,...,M_i\},
      \end{split}
    \end{equation}
where $F_i$ and $M_i$ denote modality-normalized features at scale $i$ ($i=0,\dots,4$ in our implementation). 

\subsection{Cross-scale Disentanglement and Adaptive Projection}
After obtaining the multi-scale features, we introduce a Cross-scale Disentanglement and Adaptive Projection (CDAP) module, which consists of three stages: \emph{Decompose}, \emph{Gate}, and \emph{Project}. The objective of CDAP is to generate clean and alignable multi-scale shared representations.
\textbf{Decompose:} At each scale $i$, we apply two feature extractors to decompose the feature into shared and private components:
    \begin{equation}
    \begin{split}
            F_i^{s} &= E_{sh}^i\left( F_i \right), \quad
            F_i^{p} = E_{pf}^i\left( F_i \right), \\
            M_i^{s} &= E_{sh}^i\left( M_i \right), \quad
            M_i^{p} = E_{pm}^i\left( M_i \right), 
      \end{split}
    \end{equation}
where $E_{sh}^{i}$ shares convolutional weights across modalities to extract modality-agnostic geometric cues, while $E_{pf/m}^{i}$ captures modality-specific appearance and noise characteristics. However, decomposition alone cannot prevent private-to-shared leakage, which motivates a subsequent gating step.\\
\textbf{Gate:} 
We adopt Inter-Layer Disentanglement Attention (ILDA) to suppress modality-private information, where ILDA leverages neighboring-scale semantics to perform cross-scale, attention-guided gating. For each scale $i$, we first construct query, key, and value triplets from the shared branch:
\begin{equation}
\begin{aligned}
{Q}_i^{s} &= W_q^{i}{G}(F_i^{s}),\\
{K}_i^{s} &= [W_{k}^{i-1}{G}(F_{i-1}^{s}),W_k^{i}{G}(F_{i}^{s}),W_k^{i+1}{G}(F_{i+1}^{s})],\\
{V}_i^{s} &= [W_v^{i-1}{G}(F_{i-1}^{s}),W_v^{i}{G}(F_{i}^{s}),W_v^{i+1}{G}(F_{i+1}^{s})],
\end{aligned}
\end{equation}
where $G$ denotes global average pooling, $W_q, W_k, W_v$ are learnable projections, and the brackets denote concatenation along the scale dimension. Cross-scale attention and the gating coefficient are computed as:
\begin{equation}
{c}_i^{s} = \sum_{j=1}^{3}\mathrm{softmax}_j(\tfrac{{Q}_i^{s}({K}_i^{s})^{\top}}{\sqrt{d}}){V}_{i,j}^{s}, \quad
{\alpha}_i^{s}=\sigma(\mathrm{MLP}({c}_i^{s})),
\end{equation}
where $d$ is the feature dimension, and $\sigma$ denotes the sigmoid function. We similarly obtain a gating coefficient $\alpha_i^{p}$ from the private branch using the same procedure.
Then, the attention-guided suppression is computed as:
\begin{equation}
\widetilde{F}_i^{s}={\alpha}_{i}^{s}\odot F_i^{s}-\gamma^{i}{\alpha}_{i}^{p}\odot F_i^{p},
\label{eq:ilda_suppress}
\end{equation}
where $\odot$ denotes channel-wise scaling and $\gamma^{i}\!\ge 0$ is a learnable scalar.
The same operations are applied to $M_i^{s}$ and $M_i^{p}$ to obtain $\widetilde M_i^{s}$.\\
\textbf{Project:} Next, we use Dynamic Shared Subspace (DSS) projection to map the shared features into a single low-dimensional subspace with approximately orthogonal bases for cross-scale consistent alignment.
DSS generates projection bases in an input-adaptive manner:
\begin{equation}
z_i^{s} = \mathrm{Concat}\big(\mathrm{GAP}(\widetilde{F}_i^{s}),\, \mathrm{GAP}(\widetilde{M}_i^{s})\big).
\end{equation}
then a small generator $Gen^{i}$ predicts an approximately orthonormal basis:
\begin{equation}
    W_i^{s} \;=\; Gen^{i}\!\left( z_i^{s} \right).
\end{equation}
We project the shared features into this subspace:
\begin{equation}
    \hat{F}_i^{s} = \widetilde{F}_i^{s}\, {W_i^{s}}^{\top}, \hat{M}_i^{s} =\widetilde{M}_i^{s}\, {W_i^{s}}^{\top}.
\end{equation}
The resulting $\hat{F}_i^{s}$ and $\hat{M}_i^{s}$ are the final multi-scale shared features, exhibiting reduced cross-modal statistical shift and strengthened cross-scale consistency.
\subsection{Hybrid Parameter Prediction Module}
After obtaining the multi-scale shared features $\{\hat{M}_0^{s}, \dots, \hat{M}_4^{s}\}$ and $\{\hat{F}_0^{s}, \dots, \hat{F}_4^{s}\}$ for the moving and fixed images, respectively, we predict the transformation parameters using a Hybrid Parameter Prediction Module (HPPM). 
HPPM is designed as a unified hybrid registration module that performs rigid and non-rigid registration within the same shared feature space and a single coarse-to-fine pipeline, in contrast to serial hybrids that attach a separate rigid stage followed by a non-rigid stage on different feature representations. In HPPM, global rigid alignment and local non-rigid refinement are tightly coupled: the rigid prediction at the coarsest scale is immediately encoded into a flow field and progressively refined by subsequent non-rigid updates, yielding a single coherent deformation field.

As shown in Fig.~\ref{framework}, HPPM processes feature pairs from coarse to fine resolutions. At the coarsest scale, $\hat{M}_4^{s}$ and $\hat{F}_4^{s}$ are concatenated and fed into a Hybrid Registration Block (HRB). The HRB begins with an initial convolution followed by $q=2$ Residual State Space Blocks (RSSB)~\cite{guo2024mambair}, producing the fused representation $f_4$. A registration head then maps $f_4$ to an incremental transformation parameter $\phi_4$, which is encoded as a coarse flow capturing the global alignment.
For the next scale, we first upsample the transformation parameter $\phi_{4}$ obtained at the previous scale and use it to warp the moving feature $\hat{M}_3^{s}$, producing $\hat{M}_3^{s\prime}$. We then concatenate $\hat{M}_3^{s\prime}$, $\hat{F}_3^{s}$, and the fused feature $f_{4}$, and feed them into a new HRB at this scale. The last RSSB outputs the fused representation $f_{3}$, which is passed to the registration head to generate the incremental transformation parameter $\phi_{3}'$. The updated transformation is obtained by accumulating the previous estimate and the increment:
\begin{equation}
\phi_{3} = \mathrm{upsample}(\phi_{4}) + \phi_{3}'.
\end{equation}
This coarse-to-fine refinement is repeated across all remaining scales until the finest level is reached, at which point we obtain the final hybrid transformation field $\phi$. In this way, each scale refines the deformation in a shared feature space that already encodes the accumulated global and local alignment estimated at coarser scales.

\noindent\textbf{Hybrid Registration Block.}
The HRB is the core unit of HPPM and employs two specialized heads for rigid and non-rigid registration on top of the fused representation. The rigid registration head maps the fused features $f_i$ at the coarsest scale to a low-dimensional parameter vector $H$ via global average pooling followed by two fully connected layers, capturing the global rigid transformation. The resulting $H$ is then encoded into a coarse deformable field, which is fed into finer scales through the accumulation process described above. For the non-rigid registration head, the fused features are directly mapped to a dense deformation field through a stack of $3\times3$ convolutional layers, enabling fine-grained local refinement.

In practice, HPPM is implemented with five scales: the coarsest scale is dedicated to estimating the global rigid transformation, while the remaining four scales progressively refine local non-rigid deformations. Thanks to the RSSB blocks, HPPM can capture long-range dependencies within each scale with low computational overhead, while the coarse-to-fine accumulation mechanism ensures that rigid and non-rigid transformations are predicted in a coordinated manner and composed into a single unified deformation field.
\subsection{Loss Functions}  
The overall loss consists of registration losses and disentanglement losses: the former optimizes rigid and non-rigid transformations, and the latter guides the learning of clean shared representations.

\noindent\textbf{Registration losses.}
The registration loss comprises three terms: $L_r$, $L_n$, and $L_s$.
$L_r$ and $L_n$ supervise rigid and non-rigid registration, respectively, by penalizing both parameter errors and image reprojection errors, while $L_s$ regularizes the smoothness of the deformation field.
The formula is calculated as follows:
 \begin{equation}
\begin{aligned}
L_{r}&=   \big\|\theta^{pre} - {\theta^{gt}}\big\|_1 + \big\|I_{r}^{rig} - I_{r}^{rig^{gt}}\big\|_1, \\
L_{n}&=   \big\|\phi^{pre} - {\phi^{gt}}\big\|_1 + \big\|I_{r}^{non} - I_{r}^{gt}\big\|_1,
\end{aligned}
\end{equation}
where $\theta^{pre}$ and $\phi^{pre}$ denote the predicted rigid parameters and final hybrid deformation field, respectively, and $\theta^{gt}$ and $\phi^{gt}$ are their ground-truth counterparts.
$I_{r}^{rig}$ and $I_{r}^{non}$ denote the rigidly registered image and the final hybrid registered image, while $I_{r}^{rig^{gt}}$ and $I_{r}^{gt}$ are their corresponding ground-truth images.

The smoothness term $L_s$ penalizes large spatial gradients in the deformation field to avoid unrealistic, highly oscillatory warps:
\begin{equation}
L_{s}=\frac{1}{S}\sum_{p\in\phi}\!\left[\left (  \frac{\partial\phi (p) }{\partial x} \right ) ^{2}
                                      +\left (  \frac{\partial\phi (p) }{\partial y} \right ) ^{2}  \right],
\end{equation}
where $p$ denotes a spatial location in the deformation field $\phi$, and $S$ is the total number of locations.

\noindent\textbf{Disentanglement losses.}
The disentanglement loss comprises four terms:
cross-covariance decorrelation $L_{ccd}$, basis orthogonality $L_{bo}$,
cross-scale directional consistency $L_{cs}$, and triplet loss $L_{tri}$.
Together, they shape the shared feature space so that it is weakly coupled with private components, non-redundant, and cross-scale consistent, while preserving cross-modal alignment.

$L_{ccd}$ reduces statistical coupling between shared and private components so that the shared space carries as little modality-private information as possible:
\begin{equation}
\begin{aligned}
L_{ccd}=\frac{1}{L}\sum_{i=1}^{L}w^{(i)}&(\left\| \operatorname{Cov}\!\big(\hat {F}_i^{s}, {F}_i^{p}\big) \right\|_{F}^{2} + \left\| \operatorname{Cov}\!\big(\hat {M}_i^{s},  {M}_i^{p}\big) \right\|_{F}^{2}),
\\
\operatorname{Cov}(X,Y)&=\frac{1}{B_s-1}\,X^{\top}Y,
    \end{aligned}
    \end{equation}
where $w^{(i)}$ denotes the scale weight, $B_s$ is the batch size, and $\|\cdot\|_F$ denotes the Frobenius norm. 

$L_{bo}$ encourages the DSS projection bases to be approximately orthonormal, avoiding subspace degeneracy and redundancy:
\begin{equation}
L_{bo}=\frac{1}{L}\sum_{i=1}^{L}\left\| W^{(i)}{W^{(i)}}^{\!\top}-I \right\|_{F}^{2},
\end{equation}
where $W^{(i)}$ is the DSS basis at scale $i$ and $I$ is the identity matrix. $L$ represents the number of layers

$L_{cs}$ encourages the shared features at neighboring scales to point in similar semantic directions, promoting cross-scale consistency of the shared space:
\begin{equation}
L_{cs}=\frac{1}{L-1}\sum_{i=1}^{L-1}\Big(1-\cos\big(\hat F_i^{s},\hat F_{i+1}^{s}\big)\Big),
\end{equation}
where $\cos(\cdot,\cdot)$ denotes cosine similarity between features at adjacent scales.

$L_{tri}$ aligns cross-modal co-located features in the shared space while pushing away modality-private distractors:
\begin{equation}
\begin{aligned}
L_{tri} = \frac{1}{L}\sum_{i=1}^{L} \Big[
&\max\!\big(0,\; m + d( \hat{F}_i^{s}, \hat{M}_i^{s}) - d(\hat{F}_i^{s}, F_i^{p})\big) \\
+&\max\!\big(0,\; m + d( \hat{M}_i^{s}, \hat{F}_i^{s}) - d(\hat{M}_i^{s}, M_i^{p})\big)
\Big],
\end{aligned}
\end{equation}
where $m\!>\!0$ is the margin and $d(\cdot,\cdot)$ denotes a distance measure in the shared space. The final total loss is:
 \begin{equation}  
 \begin{split}
 Loss=\alpha_rL_{r}+\alpha_nL_{n}+\alpha_sL_{s}+ \alpha_{tri}L_{tri}+\\\alpha_{cs}L_{cs}+\alpha_{ccd}L_{ccd}+\alpha_{bo}L_{bo},
 \end{split}
    \end{equation}
where $\alpha_r$, $\alpha_n$, $\alpha_s$, $\alpha_{tri}$, $\alpha_{cs}$, $\alpha_{ccd}$, and $\alpha_{bo}$ denote the loss weights.   

\section{Experiments}
\subsection{Datasets}
We evaluate HRNet on UAV-captured RGB–Thermal (TIR) and RGB–near-infrared (NIR) datasets for natural scenes, as well as on remote-sensing RGB–infrared (IR) and RGB–SAR datasets.
For RGB-TIR task, we utilize the TBBR dataset ~\cite{yasuma2010generalized}, sampling 3,000 pairs for training and 300 pairs for testing.
For RGB-NIR task, we use data from the RGB-NIR Scene dataset~\cite{brown2011multi}, which includes 442 large-size image pairs. We crop these images to generate 3,000 pairs for training and 300 pairs for testing.
For RGB-IR and RGB-SAR tasks, we employ the MRSR dataset~\cite{xiao2024adrnet}, consisting of 3,850 pairs of RGB-SAR images and 4,000 pairs of RGB-IR images. For both datasets, we use 3,500 pairs for training and the remaining data for testing. 


\subsection{Implementation Details}

Our model is implemented in PyTorch and trained on an NVIDIA L40 GPU. We use the Adam optimizer~\cite{kingma2014adam} with a learning rate of 1e\text{-}4 and a batch size of 8 for 100 epochs. All images are resized to $256\times256$ pixels.
For the rigid registration task, we generate training data with in-plane rotations in the range of $[-20^\circ, 20^\circ]$, translations in $[-15\%, 15\%]$, and scaling in $[-13\%, 13\%]$. Based on the rigid transformation, we further apply a non-rigid transformation with a deformation degree of 140 and a Gaussian smoothing radius of 35 for the non-rigid registration task.
We adopt a three-phase curriculum training strategy, adjusting the loss weights across warmup, mid, and late stages to progressively refine alignment performance, as summarized in Table~\ref{loss1}. Following~\cite{xiao2024adrnet, xiao2025multi}, we use reprojection error (RE) and normalized correlation coefficient (NCC) to evaluate both rigid and non-rigid registration.

\begin{table}[t!]
\centering
\caption{Loss weight configuration across three-phase curriculum training.}
\resizebox{1\columnwidth}{!}{
\renewcommand{\arraystretch}{1.1}{
\begin{tabular}{ccccccccc}
\hline
 stage  &  & $\alpha_r$ & $\alpha_n$  & $\alpha_s$ & $\alpha_{tri}$ & $\alpha_{cs}$ &$\alpha_{ccd}$  &$\alpha_{bo}$ \\ \cline{1-1} \cline{3-9} 
 warmup (10\%) &  & 7 & 6 & 0.5 & 0.5 & 0.05 & 0 & 0.1 \\
 mid (50\%)   &  & 5 & 10 & 0.5 & 1 & 0.1 & 0.05 & 0.2 \\
 late (40\%)   &  & 3 & 12 & 0.7 & 1 & 0.1 & 0.05 & 0.2 \\ \hline
\end{tabular}}}
\label{loss1}
\end{table}

\subsection{Results and Analysis}
For rigid registration, we compare HRNet against IHN~\cite{cao2022iterative}, InMIRNet~\cite{deng2023interpretable}, RHWF~\cite{cao2023recurrent}, SCPNet~\cite{zhang2024scpnet}, MCNet~\cite{zhu2024mcnet}, and MMRNet~\cite{xiao2025multi}.
For non-rigid registration, we compare HRNet against SuperFusion~\cite{tang2022superfusion}, InMIRNet~\cite{deng2023interpretable}, MMRNet, NBRNet~\cite{xu2022nbr}, and ADRNet~\cite{xiao2024adrnet}.

\subsubsection{Quantitative and Qualitative comparisons}
Table~\ref{rigid} and Table~\ref{nonrigid} report quantitative results for rigid and non-rigid registration, respectively.
For the rigid task, HRNet consistently achieves the best performance on all four modality pairs, with the lowest RE and highest NCC.
Compared with a strong baseline, MMRNet, HRNet reduces RE by 75.3\%, 69.9\%, 86.9\%, and 55.3\% on RGB–NIR, RGB–TIR, RGB–IR, and RGB–SAR, respectively (about 72\% relative improvement on average), while also yielding higher NCC in all cases.
For example, on RGB–IR, HRNet attains an RE of 0.578 (vs. 4.406 for MMRNet) and an NCC of 0.9172, indicating excellent pixel-level consistency.
Even on the challenging RGB–SAR dataset, HRNet still provides a clear margin, reducing RE from 7.075 to 3.161 and increasing NCC from 0.7419 to 0.8664.

For the non-rigid task, HRNet likewise shows consistent superiority, outperforming all existing methods on both RE and NCC across all modality pairs.
Relative to the previous hybrid registration method ADRNet, HRNet reduces RE by 61.2\%, 62.5\%, 66.9\%, and 23.3\% on RGB–NIR, RGB–TIR, RGB–IR, and RGB–SAR, respectively.
Compared with the strongest non-hybrid baseline MMRNet, the RE reductions are 56.7\%, 23.4\%, 73.9\%, and 17.0\% on the same four datasets.
These results demonstrate that the proposed joint rigid–non-rigid modeling substantially enhances multimodal registration accuracy and robustness, overcoming the limitations of single-paradigm deformation models.
\begin{table}[t!]
\centering
\Huge
\caption{Registration results ($mean_{std}$) for the rigid registration task. The best results are marked in \textbf{bold} and second bests are \underline{underlined}.}
\resizebox{1.01\columnwidth}{!}{
\renewcommand{\arraystretch}{1.5}{
\begin{tabular}{c|cc|cc|cc|cc}
\hline
\multirow{2}{*}{Methods}    & \multicolumn{2}{c|}{RGB-NIR}    & \multicolumn{2}{c|}{RGB-TIR}   & \multicolumn{2}{c|}{RGB-IR} & \multicolumn{2}{c}{RGB-SAR}         \\ \cline{2-9} 
                                                                           &RE$\downarrow$  &NCC$\uparrow$            &RE$\downarrow$  &NCC$\uparrow$         &RE$\downarrow$  &NCC$\uparrow$         &RE$\downarrow$  &NCC$\uparrow$   \\ \hline
IHN       & $3.887_{0.009}$ &   $0.656_{0.000}$    & $3.006_{0.004}$ &   $0.695_{0.000}$    & $5.684_{0.004}$  &   $0.747_{0.000}$    & $7.087_{0.030}$  &   $0.741_{0.003}$       \\
InMIRNet   & $7.312_{0.018}$ &   $0.509_{0.001}$    & $5.952_{0.053}$ &   $0.558_{0.003}$    & $17.958_{0.158}$ &   $0.268_{0.009}$    & $15.749_{0.223}$ &   $0.440_{0.012}$     \\
RHWF       & $5.081_{0.011}$ &   $0.616_{0.001}$    & $3.994_{0.024}$ &   $0.654_{0.001}$    & $7.289_{0.033}$  &   $0.692_{0.001}$    & $8.446_{0.006}$  &   $0.697_{0.000}$      \\
SCPNet     & \underline{$3.056_{0.037}$} &   \underline{$0.692_{0.002}$}    & $3.187_{0.279}$   &   $0.674_{0.019}$    & $6.156_{0.378}$  &   $0.722_{0.017}$    & $16.974_{0.396}$ &   $0.376_{0.012}$       \\
MCNet      & $3.582_{0.012}$ &   $0.666_{0.001}$    & $3.221_{0.023}$ &   $0.678_{0.001}$    & $12.281_{0.275}$ &   $0.483_{0.011}$    & $8.843_{0.147}$  &   $0.682_{0.005}$       \\
MMRNet     & $3.179_{0.130}$ &   $0.684_{0.009}$    & \underline{$2.472_{0.091}$} &  \underline{$0.7132_{0.005}$}    & \underline{$4.406_{0.315}$}  &  \underline{$0.799_{0.014}$}    & \underline{$7.075_{0.094}$}  &  \underline{$0.742_{0.003}$}       \\ \hline
Ours      & $\mathbf{0.785}_{\mathbf{0.013}}$
& $\mathbf{0.788}_{\mathbf{0.001}}$
& $\mathbf{0.744}_{\mathbf{0.053}}$
& $\mathbf{0.796}_{\mathbf{0.002}}$
& $\mathbf{0.578}_{\mathbf{0.019}}$
& $\mathbf{0.917}_{\mathbf{0.000}}$
& $\mathbf{3.161}_{\mathbf{0.152}}$
& $\mathbf{0.866}_{\mathbf{0.004}}$    \\ \hline
\end{tabular}}}
\label{rigid}
\end{table}

\begin{figure}[tbp!]
        \centering
	\includegraphics[scale=0.695]{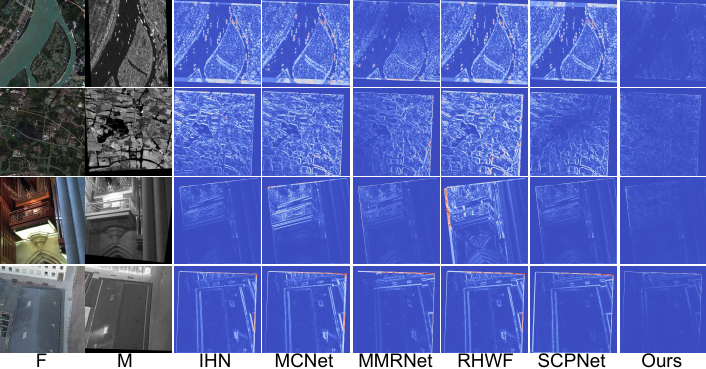}
	\caption{Qualitative comparison of rigid registration. F: fixed image, M: moving image. From top to bottom: RGB-SAR, RGB-IR, RGB-NIR, and RGB-TIR.}
        \label{vision11}
    \end{figure}

\begin{figure}[tbp!]
        \centering
	\includegraphics[scale=0.695]{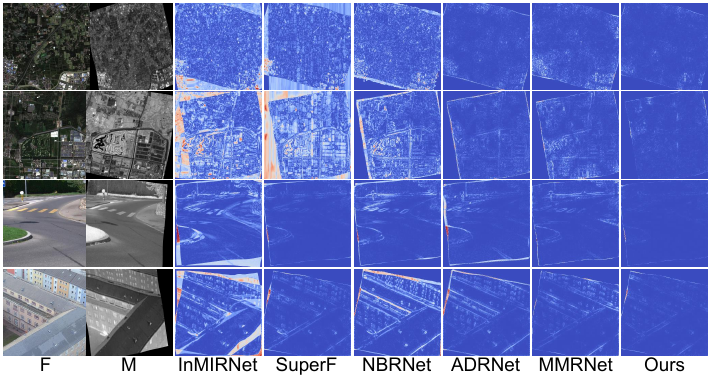}
	\caption{Qualitative comparison of non-rigid registration.  F: fixed image, M: moving image. From top to bottom: RGB-SAR, RGB-IR, RGB-NIR, and RGB-TIR.}
        \label{vision22}
        \end{figure}

\begin{table}[t!]
\centering
\Huge
\caption{Registration results ($mean_{std}$) for the non-rigid registration task. The best results are marked in \textbf{bold} and second bests are \underline{underlined}.}
\resizebox{1.01\columnwidth}{!}{
\renewcommand{\arraystretch}{1.6}{

\begin{tabular}{c|cc|cc|cc|cc}
\hline
\multirow{2}{*}{Methods}  & \multicolumn{2}{c|}{RGB-NIR} & \multicolumn{2}{c|}{RGB-TIR} & \multicolumn{2}{c|}{RGB-IR}  & \multicolumn{2}{c}{RGB-SAR} \\ \cline{2-9} 
                          &RE$\downarrow$  &NCC$\uparrow$            &RE$\downarrow$  &NCC$\uparrow$         &RE$\downarrow$  &NCC$\uparrow$         &RE$\downarrow$  &NCC$\uparrow$           \\ \hline
SuperFusion               & $4.034_{0.021}$  & $0.639_{0.001}$     & $4.325_{0.042}$  & $0.571_{0.002}$     & $36.184_{0.019}$ & $0.100_{0.000}$     & $39.341_{0.011}$ & $0.096_{0.000}$ \\
InMIRNet                  & $16.536_{0.048}$ & $0.212_{0.001}$     & $14.372_{0.055}$ & $0.211_{0.000}$     & $26.301_{0.004}$ & $0.120_{0.000}$     & $28.886_{0.046}$ & $0.110_{0.002}$   \\
NBRNet                    & $10.426_{0.110}$ & $0.351_{0.004}$     & $9.661_{0.107}$  & $0.341_{0.003}$     & $17.609_{0.146}$ & $0.275_{0.009}$     & $24.787_{0.242}$ &$0.185_{0.003}$  \\
ADRNet                    & $4.215_{0.029}$  & $0.646_{0.002}$     &$6.031_{0.382}$  & $0.490_{0.023}$     & \underline{$4.577_{0.073}$}  & \underline{$0.804_{0.003}$}     & $9.351_{0.054}$  & $0.645_{0.001}$     \\ 
MMRNet                    & \underline{$3.772_{0.043}$}  & \underline{$0.675_{0.002}$}     & \underline{$2.954_{0.002}$}  & \underline{$0.668_{0.000}$}     & $5.802_{0.031}$  & $0.738_{0.001}$     & \underline{$8.636_{0.096}$}  & \underline{$0.675_{0.003}$}   \\ \hline
Ours                & $\mathbf{1.633}_{\mathbf{0.044}}$  & $\mathbf{0.755}_{\mathbf{0.002}}$
& $\mathbf{2.264}_{\mathbf{0.029}}$  & $\mathbf{0.701}_{\mathbf{0.003}}$
& $\mathbf{1.516}_{\mathbf{0.031}}$  & $\mathbf{0.893}_{\mathbf{0.001}}$
& $\mathbf{7.172}_{\mathbf{0.131}}$  & $\mathbf{0.720}_{\mathbf{0.004}}$   \\ \hline
\end{tabular}}}
\label{nonrigid}
\end{table}
Qualitative comparisons are shown in Fig.~\ref{vision11} and Fig.~\ref{vision22}. For rigid registration, some methods produce reasonable results, but small residual rotation or translation still leads to visible global misalignment. For non-rigid registration, methods such as InMIRNet and SuperFusion often fail under large geometric discrepancies. In contrast, HRNet delivers the most visually consistent alignment in both cases, achieving more accurate geometry and better preserving cross-modal structures.


\subsection{Ablation Study}
\subsubsection{Module Ablation}

Table~\ref{tab:ablation_msbn_coda} reports the ablation results of MSBN and CDAP. On RGB–TIR, removing MSBN or CDAP degrades performance, increasing RE from 0.744 to 0.924 and 0.822, respectively, with slight drops in NCC. A similar pattern is observed on RGB–SAR: RE rises from 3.161 to 3.475 without MSBN and to 3.407 without CDAP. Overall, MSBN yields about 19.5\% and 9.0\% RE reduction on RGB–TIR and RGB–SAR, while CDAP contributes 9.5\% and 7.2\%, confirming that both components are important for high-quality hybrid registration.

\begin{figure}[tbp!]
\centering
\includegraphics[scale=0.47]{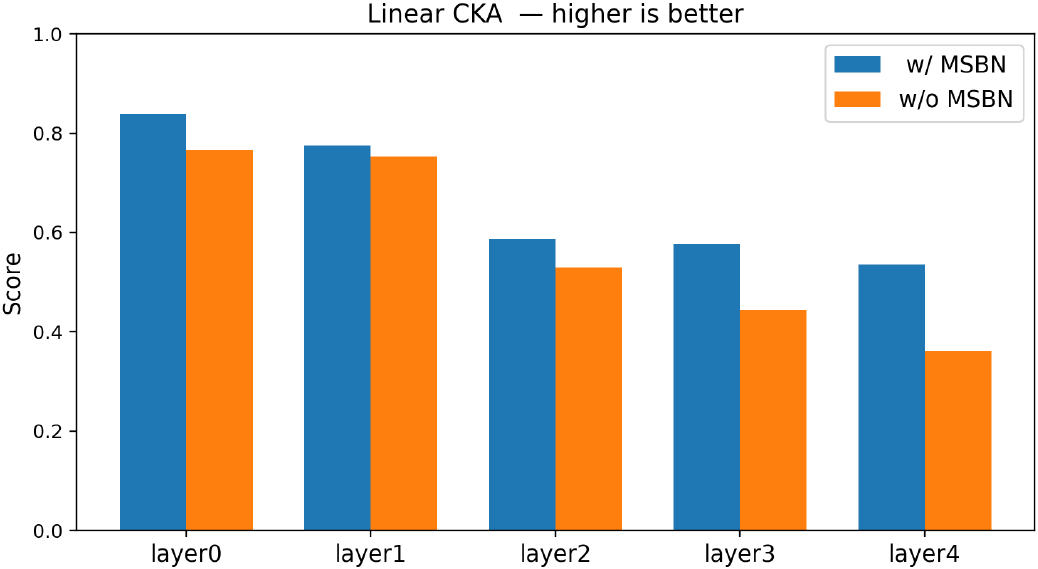}
\caption{Linear CKA comparison with/without MSBN across scales}
\label{visionMSBN}
\end{figure}

\begin{figure}[tbp!]
        \centering
    	\includegraphics[scale=1.53]{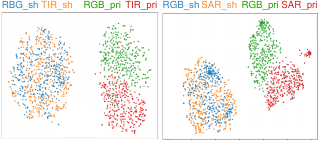}
	\caption{Visualization of disentangled feature distributions on RGB–TIR and RGB–SAR datasets.}
        \label{visioncode}
        \end{figure}

        \begin{table}[t!]
\small
\centering
\caption{Ablation of MSBN and CDAP on rigid registration for RGB–TIR and RGB–SAR datasets.}
\label{tab:ablation_msbn_coda}
\resizebox{0.7\columnwidth}{!}{
\renewcommand{\arraystretch}{0.9}{
\begin{tabular}{cccccc}
\toprule
& \multirow{2}{*}{Method} & \multicolumn{2}{c}{RGB-TIR} & \multicolumn{2}{c}{RGB-SAR} \\
\cmidrule(lr){5-6} \cmidrule(lr){3-4}
 && RE $\downarrow$ & NCC $\uparrow$ & RE $\downarrow$ & NCC $\uparrow$  \\
\midrule
(1) &w/o MSBN & 0.924 & 0.790      & 3.475 & 0.855  \\
(2) &w/o CDAP & 0.822 & 0.793      & 3.407 & 0.857 \\
(3) & Ours & $\mathbf{0.744}$ & $\mathbf{0.796}$  & $\mathbf{3.161}$ & $\mathbf{0.866}$   \\
\bottomrule
\end{tabular}}}
\end{table}  
Fig.~\ref{visionMSBN} further illustrates cross-modal consistency measured by linear centered kernel alignment (CKA) between features at the same scale across backbone layers. The model with MSBN consistently achieves higher CKA scores than its counterpart without MSBN, especially in lower layers, indicating that MSBN reduces modality-induced distribution shifts and better aligns feature geometry across modalities. This agrees with the observed gains in RE and NCC.
We also visualize feature distributions on RGB–TIR and RGB–SAR using t-SNE, as shown in Fig.~\ref{visioncode}. With CDAP, the shared components from both modalities form a single, highly overlapped cluster, while the modality-private components remain separated from each other and from the shared cluster. This suggests that CDAP suppresses private-to-shared leakage and stabilizes a registration-relevant shared subspace, enabling more reliable correspondences under large modality gaps.


\subsubsection{Loss Ablation}
Table~\ref{loss} presents a stepwise ablation of the disentanglement losses. From setting (1) to (5), progressively adding $L_{ccd}$, $L_{bo}$, $L_{cs}$, and $L_{tri}$ yields monotonic gains on both RGB–TIR and RGB–SAR: on RGB–TIR, RE decreases from 2.403 to 2.264 (-5.8\%) and NCC increases from 0.6905 to 0.7008; on RGB–SAR, RE decreases from 7.445 to 7.172 (-3.7\%) and NCC increases from 0.7125 to 0.7198. These results confirm that decorrelating shared and private components ($L_{ccd}$), enforcing an orthogonal projection basis ($L_{bo}$), aligning cross-scale semantics ($L_{cs}$), and separating shared features from private distractors ($L_{tri}$) act in a complementary way to produce cleaner shared representations and more stable multi-scale registration.

\subsubsection{Parameter Selection}

With the total number of steps fixed to 5, Table~\ref{tab:hppm_steps} shows that the setting $N_{\text{rigid}}=1$, $N_{\text{nonrigid}}=4$ consistently yields the best performance. For the rigid task, moving from an all-rigid scheme ($N_{\text{rigid}}=5$) to mixed settings steadily improves RE and NCC, and the optimum at $N_{\text{rigid}}=1$ indicates that one rigid step is sufficient for global pose correction, while multiple non-rigid steps still help refine local structures. For the non-rigid task, the same configuration also outperforms purely non-rigid ($N_{\text{rigid}}=0$) and overly rigid (e.g., $N_{\text{rigid}}\ge3$) variants, which either lack a stable global initialization or under-model fine-grained deformations. These trends support our HPPM design of using a small $N_{\text{rigid}}$ plus a larger $N_{\text{nonrigid}}$ for coherent hybrid alignment.

\begin{table}[t!]
\centering
\Large
\caption{Loss ablation on non-rigid registration for RGB–TIR and RGB–SAR datasets.}
\resizebox{0.9\columnwidth}{!}{
\renewcommand{\arraystretch}{1.1}{
\begin{tabular}{ccccccccccc}
\hline
& \multirow{2}{*}{$L_{ccd}$} & \multirow{2}{*}{$L_{bo}$} & \multirow{2}{*}{$L_{cs}$} & \multirow{2}{*}{$L_{tri}$} &  & \multicolumn{2}{c}{RGB-TIR} &  & \multicolumn{2}{c}{RGB-SAR} \\  \cline{7-8} \cline{10-11}
&                   &                        &                        &                        &  & RE $\downarrow$ & NCC $\uparrow$       &  & RE $\downarrow$ & NCC $\uparrow$        \\ \cline{1-5} \cline{7-8} \cline{10-11} 
 (1)    &             &              &             &             &  & 2.403  & 0.6905     &  & 7.445       & 0.7125          \\
 (2)    &\checkmark   &              &             &             &  & 2.356  & 0.6916     &  & 7.344       & 0.7126           \\
 (3)    &\checkmark   & \checkmark   &             &             &  & 2.309  & 0.6970     &  & 7.306       & 0.7153           \\
 (4)    &\checkmark   & \checkmark   & \checkmark  &             &  & 2.297  & 0.6982     &  & 7.242       & 0.7164           \\
 (5)    &\checkmark   & \checkmark   & \checkmark  & \checkmark  &  & 2.264  & 0.7008     &  & 7.172       & 0.7198                 \\ \hline
\end{tabular}}}
\label{loss}
\end{table}

\begin{table}[t!]
\centering
\Large
\caption{Ablation on the number of rigid vs. non-rigid steps in HPPM  on non-rigid registration for RGB–TIR and RGB–SAR datasets. (total steps fixed to 5). }
\label{tab:hppm_steps}
\resizebox{0.95\columnwidth}{!}{
\begin{tabular}{ccccccccc}
\toprule
\multirow{2}{*}{Task} & \multicolumn{2}{c}{Steps in HPPM} & & \multicolumn{2}{c}{RGB-TIR} & & \multicolumn{2}{c}{RGB-SAR} \\
\cmidrule(lr){2-3}\cmidrule(lr){5-6}\cmidrule(lr){8-9}
 & $N_{\text{rigid}}$ & $N_{\text{nonrigid}}$ & & RE $\downarrow$ & NCC $\uparrow$ & & RE $\downarrow$ & NCC $\uparrow$ \\
\midrule
\multirow{5}{*}{Rigid task}
 & 5 & 0 & & 6.378 & 0.5284 & & 15.498 & 0.4473 \\
 & 4 & 1 & & 1.872 & 0.7277 & &  9.712 & 0.6320 \\
 & 3 & 2 & & 1.344 & 0.7689 & &  7.795 & 0.7712 \\
 & 2 & 3 & & 0.801 & 0.7931 & &  4.572 & 0.8074 \\
 & \textbf{1} & \textbf{4} & & \textbf{0.744} & \textbf{0.7960} & & \textbf{3.161} & \textbf{0.8664} \\
\midrule
\multirow{5}{*}{Non-rigid task}
 & 0 & 5 & & 2.619 & 0.6743 & & 15.731 & 0.3138 \\
 & \textbf{1} & \textbf{4} & & \textbf{2.264} & \textbf{0.7008} & & \textbf{7.172} & \textbf{0.7198} \\
 & 2 & 3 & & 2.628 & 0.6730 & &  8.514 & 0.6600 \\
 & 3 & 2 & & 3.165 & 0.6273 & & 10.079 & 0.5822 \\
 & 4 & 1 & & 5.498 & 0.4495 & & 16.474 & 0.2701 \\
\bottomrule
\end{tabular}}
\end{table}
\section{Conclusion} 
This paper presents HRNet, a non-iterative coarse-to-fine multimodal hybrid registration network that jointly estimates global rigid and local non-rigid transformations within a single shared feature space. By combining a shared multi-scale backbone with modality-specific batch normalization, a cross-scale disentanglement and adaptive projection module, and a Mamba-based hybrid parameter prediction module, HRNet instantiates a generic “disentangle-then-align” template for hybrid registration. Experiments on four multimodal datasets demonstrate state-of-the-art performance on both rigid and non-rigid tasks, confirming the effectiveness of performing hybrid registration in a unified shared space. We believe this formulation provides a principled basis for extending hybrid registration to higher-dimensional, multi-view, and task-aware multimodal alignment in future work. \\
\noindent\textbf{Acknowledgments.} This work was supported by the program of China Scholarships Council under Grant 202506500011.
{
    \small
    \bibliographystyle{ieeenat_fullname}
    \bibliography{main}
}


\end{document}